\documentclass[10pt,twocolumn,letterpaper]{article}

\usepackage{style/cvpr}
\usepackage{times}
\usepackage{epsfig}
\usepackage{graphicx}
\usepackage{amsmath}
\usepackage{amssymb}

\usepackage[breaklinks=true,bookmarks=false]{hyperref}

\cvprfinalcopy 

\def\cvprPaperID{05} 

\ifcvprfinal\fi
\begin{document}

\title{TTNet: Real-time temporal and spatial video analysis of table tennis}
\author{Roman Voeikov\thanks{Authors contributed equally} \hspace{10mm}    Nikolay Falaleev$^{\ast}$ \hspace{10mm}    Ruslan Baikulov$^{\ast}$\\
OSAI\\
{\tt\small \{r.voeikov, n.falaleev, r.baikulov\}@osai.ai}
}
\maketitle
\begin{abstract}
We present a neural network TTNet aimed at real-time processing of high-resolution table tennis videos, providing both temporal (events spotting) and spatial (ball detection and semantic segmentation) data. This approach gives core information for reasoning score updates by an auto-referee system.

We also publish a multi-task dataset OpenTTGames with videos of table tennis games in 120 fps labeled with events, semantic segmentation masks, and ball coordinates for evaluation of multi-task approaches, primarily oriented on spotting of quick events and small objects tracking. TTNet demonstrated 97.0\% accuracy in game events spotting along with 2 pixels RMSE in ball detection with 97.5\% accuracy on the test part of the presented dataset.

The proposed network allows the processing of downscaled full HD videos with inference time below 6 ms per input tensor on a machine with a single consumer-grade GPU. Thus, we are contributing to the development of real-time multi-task deep learning applications and presenting approach, which is potentially capable of substituting manual data collection by sports scouts, providing support for referees' decision-making, and gathering extra information about the game process.
\end{abstract}
\section{Introduction}
\label{introduction}

\thispagestyle{empty}
Deep learning-based computer vision approaches have recently started to play an important role in sports analytics. There is a broad scope of tasks to address: tracking of players \cite{Ullah_2018_CVPR_Workshops} and sports equipment (like balls \cite{KAMBLE201958} or hockey sticks \cite{DBLP:journals/corr/abs-1812-09533}), human pose estimation \cite{Bridgeman_2019_CVPR_Workshops}, and also multiple levels of detection of game-related actions \cite{Buch_SST}. For many of these tasks, it is of high interest to achieve real-time performance to aid the on-line game analytics that demands steady and very fast solutions.

Table tennis is a fast-paced game with a great variety of visual data to be analyzed. Replacing manual data collection by automatic systems would potentially allow increasing the speed and accuracy of video analysis.

\begin{figure}[t]
\begin{center}
   \includegraphics[width=0.95\linewidth ]{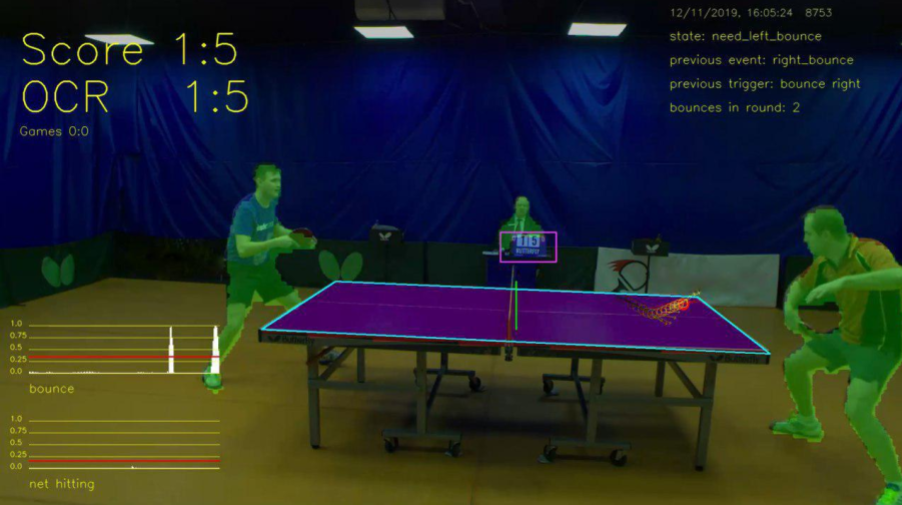}
\end{center}
   \caption{Sample frame with the TTNet predictions overlaid.}
\label{fig:scene}
\end{figure}

The essential events in table tennis, which define the game process after a serve, are ball bounces and net touches. Manual collection of information about the number of bounces and their coordinates on the table is almost impossible because of high ball speed, but it may be accomplished with deep learning-based video analysis. However, there is a broad scope of challenges associated with this task. For example, the size of the ball in a full HD video, which contains both players, an umpire and a table, is quite small: about 15 pixels on average. Moreover, the ball may be not the only small white object in the image because of the parts of player clothing or background elements. Another challenge is the ball speed. Indeed, during the intense game, the speed of the ball may be more than 30 m/s. Therefore, video data with a high frame rate is required to detect the trajectory of the ball and corresponding events, like bounces and net hits.

All this leads to many limitations and high demands on neural network architectures. First of all, the network must have low inference time to process each frame of the high frame rate video in real-time, using reasonable equipment like a machine with a single GPU. Thus, this brings us to the field of shallow convolutional networks with a relatively small number of operations. Moreover, different types of information need to be extracted from the input video in parallel: segmentation masks (humans, table and scoreboard), coordinates of the ball, and game events. In this work, we present a relatively lightweight multi-task architecture TTNet to address all these tasks simultaneously, having 5.9 GFLOPs and inference time below 6 ms on a machine with a single NVIDIA RTX 2080Ti GPU.

Another challenge is the lack of public multi-task datasets with the main focus on spotting quick game events. Therefore, we had to introduce a dataset for the assessment of the proposed architecture. To address this issue, we collected and labeled table tennis videos. In addition to the model architecture, we release the dataset for the community, to enable consistent assessment of computer vision techniques for table tennis analysis.
\section{Related works}
\label{realted works}

Despite the table tennis popularity, only a very limited number of works offer computer vision-based solutions for the tasks related to this sport. Previous works were mainly focused on the ball detection. For example, in the work of Tamaki and Saito \cite{Tamaki_Saito_3D_traj} contour analysis methods were used to find potential ball candidates. However, such detection approach faces problems with ball-like objects, generating many false positive detections.

Similarly, in the project of Myint \etal \cite{Myint_Tracking_2015} adaptive color thresholding and background subtraction are used for the ball segmentation on stereo-images. In the final setup, presented in \cite{Myint_Tracking_2016}, a system of 4 interrelated high-speed cameras was involved for the reconstruction of the ball 3D trajectory that leads to a very accurate ball detection but requires complex equipment.

Besides table tennis, some works investigated the capabilities of Deep Learning solutions for a ball tracking in other sports. Reno \etal \cite{Reno_2018_CVPR_Workshops} proposed the CNN architecture for detecting the ball in videos of tennis games by classifying small patches of the input image to decide if it contains the ball or not. However, the treatment of a significant amount of overlapping patches is not suitable for real-time applications.
Komorowski \etal \cite{DBLP:journals/corr/abs-1902-07304} presented novel CNN architecture DeepBall, which was designed to find the ball during football games. It works in 190 fps on full HD images, achieving state-of-the-art performance in this task when the ball moves relatively slow, while on table tennis videos the ball speed is much higher.

However, the main goal of the solutions above was only the ball detection, while sports videos contain much more useful information about players, game events and environmental conditions, which may be exploited for a more comprehensive analysis of players' performance, for development of auto-umpire systems, and for the building of predictive machine learning models. Multi-task models (\cite{Sermanet_OverFeat}) may be used to deal with the mentioned tasks, simultaneously outputting different types of information. For instance, UberNet \cite{Kokkinos_2017_CVPR} was proposed as an architecture aimed at solving a broad scope of computer vision problems: object detection, human parts segmentation, etc. All tasks were addressed by combining different datasets and training this neural network in an end-to-end manner, which significantly simplifies the training process, making it more clear and straightforward.

Also, the multi-task approach was used to enhance action recognition accuracy in the work of Luvizon \etal, where pose estimation was combined with visual features helps to classify actions in video sequences \cite{DBLP:journals/corr/abs-1802-09232}. However, to the best of our knowledge, the problem of tracking very small objects in the image, combined with a multiclass semantic segmentation and temporal events spotting, has not been addressed in a single CNN architecture yet that brought us to the development of TTNet.
\begin{figure*}
\begin{center}
\includegraphics[width=0.9\linewidth]{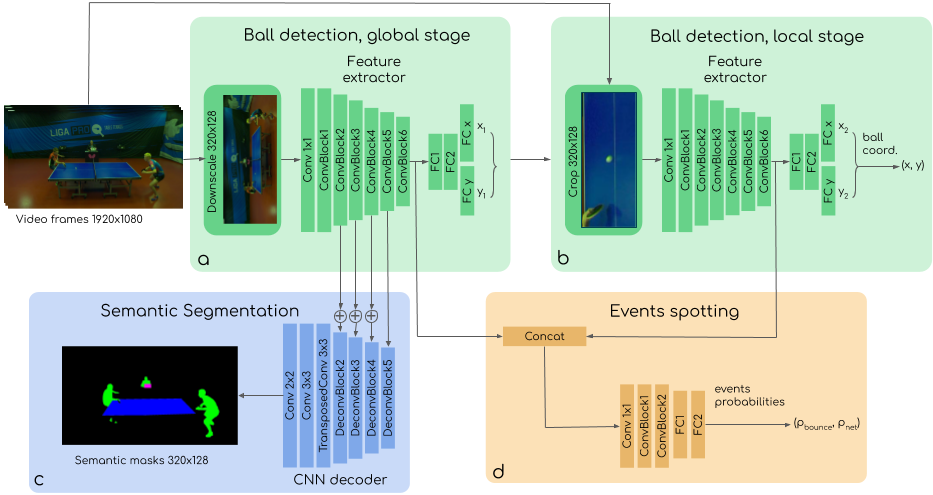}
\end{center}
   \caption{TTNet architecture: a and b - ball detectors, a - works on downscaled frames, b - on crops from full HD; c - semantic segmentation tail resulting in table, scoreboard and human masks; d - events (bounces and net hits) spotting tail; $\bigoplus$ operation denotes summation of feature maps from corresponding encoder layer and from previous decoder layer.}
\label{fig:TTNet}
\end{figure*}

\section{Table tennis analysis}
\label{table tennis analysis}
The reasoning of score updates during the rally by an auto-referee system requires different types of information to be gathered. The core of the rules is in-game events: ball bounces and net hits, the correct treatment of which gives an understanding of the current game state. However, the coordinates of the ball and the position of the table are necessary to determine the bounce side as well as to filter out false-positive bounces that occur not on the table. Moreover, the table position may be slightly changed during the game, so it needs to be updated continuously. Player positions are also required to prevent the extraction of information about the table when it may be partially occluded, which can lead to an incorrect determination of the net position. Also, the mask of the referee could be utilized for the detection of the gesture of let serve events. Finally, the scoreboard position is also essential because a parallel system can use it for the OCR retrieval of the ground truth score.

Taking into account all the required data, we developed an architecture that simultaneously solves the tasks of event spotting, object detection, and semantic segmentation.

\section{OpenTTGames Dataset}
\label{OpenTTGames dataset}

Due to absence of multi-task datasets combining temporal and spatial data we introduce OpenTTGames\footnote{\nolinkurl{https://lab.osai.ai/datasets/openttgames}}, covering the mentioned aspects of table tennis analysis, which was used for carrying out experiments in this paper. It includes full HD videos of table tennis games recorded at 120 fps. Every video is equipped with an annotation, containing the frame numbers and corresponding targets for this particular frame: manually labeled in-game events (ball bounces, net hits, or empty event targets needed for the CNN robustness and as a tool to avoid overfitting) and/or ball coordinates and segmentation masks, which were labeled with deep learning-aided annotation models. There are 5 videos of different matches, intended for training, and 7 short videos from other matches for testing. Every video is recorded in similar conditions with slight variations in camera angle. Taking into account our method of target construction, described in \ref{data preparation}, it has resulted in 38752 training, 9502 validation, and 7328 testing samples.

\section{Proposed method}
\label{proposed method}

The whole architecture of TTNet is depicted in Fig.  \ref{fig:TTNet}, while its main building blocks are described in Tab. \ref{tab:ConvBlock} and Tab. \ref{tab:DeconvBlock}. The network consumes input tensors, formed by stacks of 9 subsequent frames from a raw video (resolution: \(w_0 = 1920 px; h_0 = 1080 px)\) sampled at 120 fps, and outputs in-game events probabilities (bounces on the table and the net hits), semantic masks of the table, humans, and scoreboard as well as the ball position. The downscaled input is processed by the core backbone - VGG-style feature extractor. The resulting feature maps are used in an encoder-decoder semantic segmentation branch and for the initial ball detection. The ball position estimation is rough on this stage due to low resolution. Therefore, the second stage of the ball detection is introduced. It works in the same manner but utilizes frames from the original input cropped around the prediction of the first detector. Thus, the final ball detection is performed with the spatial resolution of the original video. The events are ball-related, so it can be expected that local features in the ball area could be beneficial for the recognition of the events. For this reason, the local stage feature maps are shared with the event spotting branch. Details of the architecture are described in the following sections.

\subsection{Ball detection}
Ball positions should be estimated as precise as possible, given the camera resolution. This task involves several difficulties. Our approach assumes only one real ball on the frame, but there could be other small white round objects, which could be mistaken for the ball. In addition, the ball is usually blurred due to motion. This makes the distinctive dynamic features of the true ball motion substantial for its reliable detection. Therefore, the proposed approach uses a stack of consequent video frames rather than a single frame to capture the motion information.

\begin{figure}[t]
\begin{center}
   \includegraphics[width=1\linewidth]{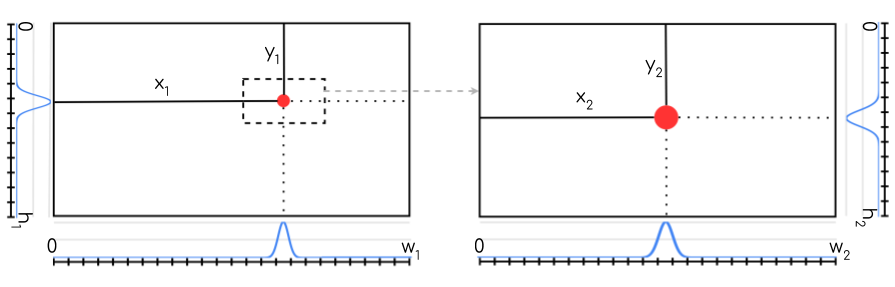}
\end{center}
   \caption{Target construction for two stages of the ball detection.}
\label{fig: ball_target}
\end{figure}

The ball localization is performed by predicting its center on the last (the most recent) frame in the stack. The target values are composed of two vectors with the length of the width and the height of the input. The values of the target vectors are produced by a normal distribution fitted around the ball center coordinates from the ground truth labels data as proposed in \cite{RoboCup}. In other words, the target (Fig. \ref{fig: ball_target}) is two one-dimensional Gaussian distribution curves with means associated with the x and y of the ball center, respectively. The variances of the curves are set to reasonable values associated with the average ball radius at the scale of the fed frames.

\begin{table}
\begin{center}
\begin{tabular}{p{28mm}|c |c |c}
\hline
Operator & \begin{tabular}{@{}c@{}}In-channels/ \\ Out-channels\end{tabular} & stride & padding \\
\hline\hline
Conv 3x3 & a/b & 1 & 1 \\
BatchNorm & b/b & - & - \\
ReLU & b/b & - & - \\
MaxPool 2x2 & b/b & 2 & 0 \\
\hline
\end{tabular}
\end{center}
\caption{Structure of the ConvBlock, "a" and "b" are arbitrary parameters}
\label{tab:ConvBlock}
\end{table}

\begin{table}
\begin{center}
\begin{tabular}{p{28mm}|c |c |c}
\hline
Operator & \begin{tabular}{@{}c@{}}In-channels/ \\ Out-channels\end{tabular} & stride & padding \\
\hline\hline
Conv 1x1 & a/\(\frac{a}{4}\) & 1 & 0  \\
BatchNorm & \(\frac{a}{4}\)/\(\frac{a}{4}\) & - & - \\
ReLU & \(\frac{a}{4}\)/\(\frac{a}{4}\) & - & - \\
TConv 3x3, op=1 & \(\frac{a}{4}\)/\(\frac{a}{4}\) & 2 & 1 \\
BatchNorm & \(\frac{a}{4}\)/\(\frac{a}{4}\) & - & -  \\
ReLU & \(\frac{a}{4}\)/\(\frac{a}{4}\) & - & -  \\
Conv 1x1 & \(\frac{a}{4}\)/b & 1 & 0 \\
BatchNorm & b/b & - & - \\
ReLU & b/b & - & - \\
\hline
\end{tabular}
\end{center}
\caption{Structure of the DeconvBlock. "TConv" refers to Transposed Convolution, "op" refers to output padding, "a" and "b" are arbitrary parameters}
\label{tab:DeconvBlock}
\end{table}

\begin{table}
\begin{center}
\begin{tabular}{p{18mm}| c|c}
\hline
Input size & Operator & \begin{tabular}{@{}c@{}}In-channels/ \\ Out-channels\end{tabular} \\
\hline\hline
320 x 128 & Conv 1x1 & 27/64 \\
320 x 128 & BatchNorm & 64/64 \\
320 x 128 & ReLU & 64/64 \\
320 x 128 & ConvBlock & 64/64 \\
160 x 64 & ConvBlock & 64/64 \\
80 x 32 & DropOut & 64/64 \\
80 x 32 & ConvBlock & 64/128 \\
40 x 16 & ConvBlock & 128/128 \\
20 x 8 & DropOut & 128/128 \\
20 x 8 & ConvBlock & 128/256 \\
10 x 4 & ConvBlock & 256/256 \\
5 x 2 & DropOut & 256/256 \\
5 x 2 & Flatten & 256/- \\
2560 & FC & - \\
1792 & ReLU & - \\
1792 & DropOut & - \\
\hline
1792 & FC & - \\
640/256 & ReLU & - \\
640/256 & DropOut & - \\
320/128 & FC & - \\
320/128 & Sigmoid & - \\
\hline
\end{tabular}
\end{center}
\caption{Structure of the Ball Detection part (global and local structures are identical). There are 2 parallel sets of operators after the horizontal line, giving outputs for x and y vectors respectively}
\label{tab:BallDetector}
\end{table}

The ball detection is done on two scales. The first detection stage (referred to as global) is made on a downscaled video frames (to size \(w_1 = 320 px; h_1 = 128 px)\). The usage of the reduced input provides fast localization of the ball on the whole frame. The input scale for this step is chosen to ensure the ball to be at least two pixels in diameter. The second stage ball localization (referred to as local) is accomplished on crops (crop size: \(w_2 = 320 px; h_2 = 128 px\)) from the original video frames around the detected center of the ball by the first stage. Both stages of the ball localization share the same architecture of the neural network and the same way of the target construction. The final ball coordinates are derived from the two stages results by Eq.\ref{eq:xy_ball}.

\begin{equation}\label{eq:xy_ball}
x = x_1 \frac{w_0}{w_1} - \frac{w_2}{2} + x_2;  \quad  y = y_1 \frac{h_0}{h_1} - \frac{h_2}{2} + y_2
\end{equation}

The network architecture involved in this task consists of a convolutional encoder (feature extractor) and a fully connected tail. The results of the convolutional backbones from both detectors are shared with other task branches. The features, which were produced while processing downscaled images of the whole scene, are used for semantic segmentation, whereas the features based on the cropped frames are utilized for the event spotting task. The encoder is made up of the 1x1 Convolutional layer, followed by six Convolutional blocks (Tab. \ref{tab:ConvBlock}) wired sequentially. The following fully connected part is formed in a swallow-tail shape, as the neural network has to predict a pair of coordinate vectors (Tab. \ref{tab:BallDetector}).

The branch is trained with a cross-entropy loss function, calculated for x and y vectors independently and summed up on both scales (Eq. \ref{eq:Loss_ball}):

\begin{equation}\label{eq:Loss_ball}
L_{ball_{1,2}} = -\frac{1}{w_{1,2}}\sum_{i=1}^{w_{1,2}} {\hat{p}^{i}_{x_{1,2}}} \log {p}^{i}_{x_{1,2}}   -\frac{1}{h_{1,2}}\sum_{i=1}^{h_{1,2}} \hat{p}^{i}_{y_{1,2}} \log {p}^{i}_{y_{1,2}}
\end{equation}

, where \(\hat{p}^{i}\) - target vector values, \({p}^{i}\) - predicted values.

\subsection{Event spotting}

\begin{table}
\begin{center}
\begin{tabular}{c|c|c}
\hline
Input size & Operator & \begin{tabular}{@{}c@{}}In-channels/ \\ Out-channels\end{tabular} \\
\hline\hline
5 x 2 & Conv 1x1 & 512/64 \\
5 x 2 & BatchNorm & 64/64 \\
5 x 2 & ReLU & 64/64 \\
5 x 2 & DropOut & 64/64 \\
5 x 2 & ConvBlock (w/o MaxPool) & 64/64 \\
5 x 2 & DropOut & 64/64 \\
5 x 2 & ConvBlock (w/o MaxPool) & 64/64 \\
5 x 2 & DropOut & 64/64 \\
5 x 2 & Flatten & 64/- \\
640 & FC & - \\
512 & ReLU & - \\
512 & FC & - \\
2 & Sigmoid & - \\
\hline
\end{tabular}
\end{center}
\caption{Structure of the Events Spotting part}
\label{tab:EventTail}
\end{table}

According to the table tennis rules, only ball bounces, serves, and net hits are essential to keep the score updated. Due to the fact that the neural network feed is a 9 frame sequence, which timespan is less than 0.1 s, the architecture is capable of detecting only fast events, such as bounces and net hits. Thus, the TTNet model is used for the auto-referee purposes only during the rally, while serves, indicating the start of the rally, could be recognized by a similar network. It works with temporarily wider frame sequences because serves have a great variety of possible duration; however, this task is out of the scope of the paper.

The event spotting branch acts on concatenated feature maps from global and local detectors. Such an approach helps gradients from event spotting loss to flow through both feature extractors. The branch consists of three convolutional blocks without MaxPool followed by two fully connected layers (Tab. \ref{tab:EventTail}).

The last activation layer (Sigmoid) allows both events to occur simultaneously. Thus, the event spotting task may be considered as a multilabel classification. This is motivated by the fact that binary labels are not practically suitable for this task due to the high speed of the ball. It may be impossible to pick the particular frame with the event because the ball may be blurred in motion, or there may not be a frame with an actual bounce at all. This leads to the necessity of smooth event labeling. The target values were constructed as \(\sin{\frac{n\pi}{8}}\) to be in (0,1) range and act as events probabilities, where n is the number of frames between the considered frame and the manually labeled event frame. The target is nonzero if \( n \in (-4, 4) \)  and 0 otherwise.

The events of the considered types could not be spotted without information on the ball trajectory after the potential event moment. The only input of the system is 2D images without depth data; hence, one can not say if the ball is already on the table surface or it will continue its dive on the next frame. For this reason, the central frame in the image stack is labeled with appropriate target probabilities. This way of the target construction allows including data from the past and the future frames with respect to the assessed one to spot the events reliably. This leads to a certain delay in the prediction; however, in our case, it is under 0.05 s, which is far below the reaction time for humans. Consequently, it does not reduce the subjective quality of predictions on live video, and it is not practically significant.

Because the problem is reduced to classification, the event spotting branch was trained with a weighted (1:3 bounce:net hit) cross-entropy loss (Eq. \ref{eq:Loss_event}) due to a significant class imbalance. Randomly sampled frames sequences without events were added to the training dataset with zero targets to make the neural network correctly predict the event absence. A number of these negative samples was roughly the same as the number of labeled events.

\begin{equation}\label{eq:Loss_event}
L_{event} =-\frac{1}{N_{events}}\sum_{i \in \{events\}} \beta_{i} \hat{p}^{i} \log {p}^{i}
\end{equation}
, where \(N_{events}\) - number of possible event types (2 in the case), the summation is performed over these event types.

\subsection{Semantic segmentation}

\begin{table}
\begin{center}
\begin{tabular}{p{18mm}| c|c}
\hline
Input size & Operator & \begin{tabular}{@{}c@{}}In-channels/ \\ Out-channels\end{tabular} \\
\hline\hline
10 x 4 & DeconvBlock & 256/128 \\
20 x 8 & DeconvBlock & 128/128 \\
40 x 16 & DeconvBlock & 128/64 \\
80 x 32 & DeconvBlock & 64/64 \\
160 x 64 & TConv 3x3, s=2, p=0, op=0 & 64/32 \\
321 x 129 & ReLU & 32/32 \\
321 x 129 & Conv 3x3, s=2, p=0 & 32/32 \\
319 x 127 & ReLU & 32/32 \\
319 x 127 & Conv 2x2, s=2, p=1 & 32/3 \\
320 x 128 & Sigmoid & 3/3 \\
\hline
\end{tabular}
\end{center}
\caption{Structure of the Semantic Segmentation Decoder part. "TConv" - Transposed Convolution, "op" - output padding, "p" - padding, "s" - stride}
\label{tab:SegmTail}
\end{table}

The required semantic masks of three classes (humans, table, scoreboard) are predicted in the last frame of the stack using an encoder-decoder approach. The encoder is the backbone, which is shared with the first stage ball detector. Although the masks are predicted on the downscaled input only, the resolution is enough for the practical applications. Decoder blocks (Tab. \ref{tab:DeconvBlock}) are based on 2D Transposed Convolutional layers for upsampling by a factor of 2 as opposed to the downscaling with the Max Pooling layer in the encoder blocks. The Transposed Convolutional layers are surrounded with 1x1 Convolutional layers, reducing computation by decreasing the treated tensor depth and aligning feature maps sizes on the encoder and decoder stages in order to apply skip connections.

Skip connections wiring scheme is applied in order to preserve spatial resolution. Features produced at three smallest scales before the final encoder block are bypassed into the corresponding scale decoder blocks and added-up in an element-wise manner. This approach is adopted from the LinkNet \cite{DBLP:journals/corr/ChaurasiaC17}.

The last activation layer is chosen to be a Sigmoid layer as semantic masks of different classes may be overlapped. The branch is trained with two loss functions (Eq. \ref{eq:Loss_segm}) summed up: 1 - S\o rensen-Dice coefficient (smoothed with \(\epsilon= 10^{-4} \) for numerical stability, Eq. \ref{eq:DICE_smooth}) as the first component of the loss and binary cross-entropy loss (BCE) as the second one.

\begin{equation}\label{eq:Loss_segm}
L_{segm} = (1 - DICE_{smooth}) + BCE
\end{equation}

\begin{equation}\label{eq:DICE_smooth}
DICE_{smooth} = \frac{2|\hat{P} \cap {P}| + \epsilon} {|\hat{P}| + |{P}| + \epsilon}
\end{equation}
, where \(\hat{P}\) - target binary mask, \(P\) - semantic segmentation prediction map

\subsection{Multi-task loss function}
Multi-task training requires loss aggregation. A naive weighted sum of losses for simultaneous learning of multiple tasks is widely used. The applied weights are uniform or manually tuned. The proposed neural network predicts data of different modalities, and the predictions are interconnected. For instance, the success of the event spotting branch is related to the success of the ball detection, as the events are predicted for frame crops around the proposed ball position. Moreover, the differences in the target data types lead to inconsistent learning paces. In order to overcome these issues, an approach to weight losses, considering the homoscedastic uncertainty of each task, was adopted from \cite{kendall2017multi}. It incorporates the relative weights of the losses adaptively and treats them as trainable parameters (Eq. \ref{eq:Loss}). The last term of the sum acts as a regularizer for the trivial solution elimination.

\begin{equation}\label{eq:Loss}
L = \sum _{i=1}^4 \frac{L_i}{\sigma_i^2}+ \sum _{i=1}^4 \log {(\sigma _i)}
\end{equation}
, where \(L_i\) - one of four: \(L_{ball_{1,2}}\), \(L_{events}\) or \(L_{segm}\) .

\subsection{Data preparation}
\label{data preparation}
\begin{figure}[t]
\begin{center}
   \includegraphics[width=1\linewidth]{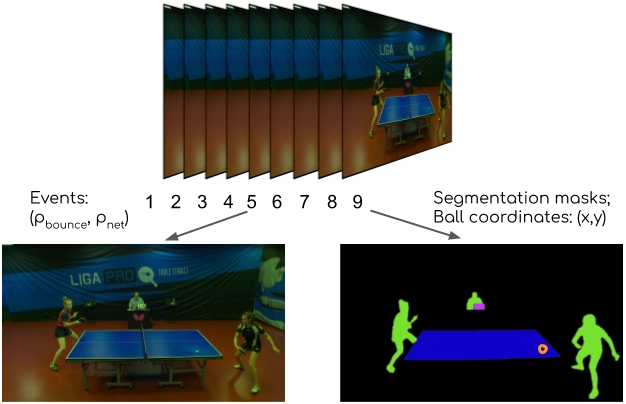}
\end{center}
   \caption{Input tensor structure. Event target is constructed for the middle (5th) frame, ball and segmentation targets - for the last (9th).}
\label{fig:input_tensor}
\end{figure}

The supervised approach requires labeled data for all three tasks. The manual event labeling is quite fast with an in-house markup tool, which allows marking frames with the events of interest. The event spotting task is considered as the key task; it is done completely manually for the whole dataset. The rest of the targets are built around it.

The training sample is represented as a sequence of 25 video frames with the manually labeled event right in the middle frame. The length of the sample allows to subsample training sequences of 9 subsequent frames to provide different stages of the considered event. Input tensors for the training were constructed from this 9 frame subsamples with event spotting target for the middle frame and semantic segmentation masks and the ball position for the last frame in such sequence (Fig. \ref{fig:input_tensor}).

The semi-automated process was utilized to facilitate the manual annotation procedure for the ball coordinates and the segmentation masks. Individual networks were used for every auto-annotation task. The networks' architecture mimics the segmentation branch of the proposed multi-task neural network but adopts more powerful but slower encoders: ResNet-152 for the human class and ResNet-34 for the table and the scoreboard classes. These supplementary neural networks were trained using data from different sources. For instance, open-source datasets were used for the human segmentation task: fine-annotated examples from Cityscapes Dataset \cite{Cordts2016Cityscapes}, Supervisely Person Dataset \cite{Supervisely} and COCO (detection 2017) Dataset \cite{DBLP:journals/corr/LinMBHPRDZ14}, alongside with manually annotated 50 images of real tennis players. A custom dataset of 1500 human-labeled images was used for the table segmentation task, while for the board segmentation, a synthetic dataset consisting of 5000 images distilled with several real examples was utilized. Every auto-labeling model was evaluated using intersection over union metric, and for human, table and scoreboard segmentation tasks it resulted in values of 0.964, 0.945, and 0.986, respectively.

The same aided way of data annotation was applied for the ball detection task. The two-stage ball detection part of the whole multi-task neural network pipeline was trained with 4 frame sequences as input tensors. 47500 raw frames were labeled by hand to form a dataset for this task. The following metrics were used to evaluate the model performance: accuracy of ball presence detections and RMSE error for detected ball coordinates (see Section 6.2). After the training process, the final ball detection model for auto-labeling was able to detect the ball in 96.3\% of cases with an average error of 2.5 pixels by the local detector and in 96.4\% of cases with an average error of 14.3 pixels by the global detector.
\section{Experiments}
\label{experiments}

\subsection{Implementation Details and Training}

As mentioned earlier, spatial and temporal multitasking is the main goal of this work, and there are no open datasets combining such data, so OpenTTGames is used for model evaluation

All experiments were carried out with PyTorch 1.2.0 framework on a PC with a single NVIDIA RTX 2080 Ti graphics card. The same setup was involved in the final pipeline performance evaluation.

During all training experiments, the Adam optimizer was used with an initial learning rate 0.001 and default parameters \(\beta _1 =0.9\), \(\beta _2 =0.999\), and \(\epsilon=10^{-8} \). The learning rate was halved after 3 epochs without the loss value decrease, whereas the finishing criterion for training was the absence of any loss decrease for 12 epochs in a row. These relatively small numbers of epochs were worked out by a manual hyperparameters tuning process.

Following image augmentations were applied: random cropping with width and height reduction up to 15\%, frames rotation (\( \pm 15 ^{\circ} \)), horizontal frames flip (which improves treatment of left-handed players), random brightness, contrast, and hue shifts. The image transformations were applied with the same parameters to each frame in a sequence in order to keep consistency.

\begin{table}
\begin{center}
\begin{tabular}{|p{32mm}|c|c|c|c|c|}
\hline
\begin{tabular}{@{}c@{}}Feature extractor\\architecture\end{tabular}  & ResNet-18 & TTNet-encoder\\
\hline\hline
Encoder GFLOPs & \textbf{2.260} & 2.340\\
Parameters, M & 11.250 & \textbf{1.180}\\
Inference time, ms & 7.6 & \textbf{6.0}\\
Global RMSE, px & 12.11 & \textbf{6.79}\\
Local RMSE, px & 1.99 & \textbf{1.97}\\
Global accuracy & 0.973 & \textbf{0.975}\\
Local accuracy & 0.973 & \textbf{0.978}\\
IoU & \textbf{0.943} & 0.928\\
PCE & 0.971 & \textbf{0.977}\\
SPCE & 0.970 & \textbf{0.970}\\
\hline
\end{tabular}
\end{center}
\caption{The TTNet VGG-style encoder (feature extractor) performs better in almost all metrics and has lower inference time (presented for the full pipeline) as compared to ResNet-18 backbone.}
\label{tab:Backbones}
\end{table}

\subsection{Evaluation metrics}

Both stages of the ball position prediction were assessed with two metrics. Initially, the predicted vectors were thresholded with a level of 0.5. Then, the first metric is intended to evaluate the accuracy of the ball presence prediction. The ball was considered to be present in the frame in case there was at least one non-zero value in both thresholded vectors. Matching these results with the ground truth data allows calculating the number of true-positive and true-negative detections, resulting in detection accuracy. The second metric was the Euclidean distance (\textbf{RMSE}) between the predicted and labeled ball position computed over true-positive ball detections only.

The event spotting branch target was designed as probability values for each event type. This makes it impossible to use sequence-based metrics, like temporal Intersection Over Union. Therefore, the Percentage of Correct Events (\textbf{PCE}) metrics was applied \cite{McNally_2019_CVPR_Workshops}. It calculates the portion of correctly spotted events. An event was considered as correct if the predicted value is equal to the ground truth after 0.5 level thresholding of the predicted and target values. It is worth noting that the metric could misclassify predictions with intermediate probabilities. A Smooth Percentage of Correct Events (\textbf{SPCE}) metric was introduced to overcome this limitation. It is the same as PCE, but an event treated as a correct one if the difference with the target is less than a threshold (0.25 in the experiments).

The semantic maps predictions were assessed with Intersection Over Union (\textbf{IoU}).

\subsection{Feature extractor}
The first object of interest in the CNN architecture was feature extractor. Due to the mentioned above reasons, it was designed to be fast, lightweight, yet powerful, with the main focus on inference time for real-time applications. We examined ResNet-18 architecture as a potential replacement for our original TTNet backbone due to its relatively small size and proven effectiveness of the general ResNet architecture in many deep learning tasks.

The results presented in Tab. \ref{tab:Backbones} demonstrate that the whole model with TTNet backbone has lower inference time, much fewer parameters, and it outperforms the model with the ResNet-18 backbone in almost every task. Training process for both encoders was conducted using the optimal number of input frames and adaptive loss balancing.
Due to our inference time limitations, we excluded from consideration many other popular architectures like DenseNet  \cite{DenseNet}, Inception  \cite{Inception}, and ResNeXt  \cite{ResNext}.

\subsection{Target sequence length}
We performed experiments to find the minimum number of subsequent frames forming an input tensor to produce the best results, without decreasing the processing speed. For this purpose, experiments with separate training of every branch of the TTNet with input tensors of different widths (from 1 to 9 subsequent frames with a step of 2) were carried out. 
In the ball detection task, the optimal numbers of frames were 3-5, demonstrating the best results on almost every metric (Tab.  \ref{tab:Len_ball}). However, the usage of 9-frame sequences resulted in just a slight results decrease in the same metrics.

\begin{table}
\begin{center}
\begin{tabular}{|p{25mm}|c|c|c|c|c|}
\hline
\begin{tabular}{@{}c@{}}Target width, \\n frames\end{tabular}  & 1 & 3 & 5 & 7 & 9 \\
\hline\hline
Global RMSE, px & 30.93 & 3.56 & \textbf{3.08} & 3.47 & 3.60 \\
Local RMSE, px & 6.64 & \textbf{1.23} & 1.36 & 1.46 & 1.46 \\
Global accuracy & 0.955 & 0.982 & \textbf{0.982} & 0.981 & 0.980 \\
Local accuracy & 0.912 & 0.983 & \textbf{0.983} & 0.983 & 0.981 \\
\hline
\end{tabular}
\end{center}
\caption{Input sequence length optimization for the ball detection part of TTNet. 3-5 frames performs better.}
\label{tab:Len_ball}
\end{table}

For the semantic segmentation task, the best results were achieved for a 1-frame sequence; however, no degradation in segmentation quality was observed in the case of longer sequences.

The event spotting branch is naturally related to the ball detector branch in TTNet, so for the experiments with the different lengths of the input sequence the network architecture included both of these branches (Tab. \ref{tab:Len_events}). The event target strongly correlated with temporal information, so the best results in event spotting and, surprisingly, in ball detection by the global detector were achieved in the case of the longest 9-frame sequences. Thus, this length was used in all further experiments because of the best results on the core task of events spotting and as a compromise with a slight degradation of accuracy in other tasks. Moreover, such length is reasonably small for providing the events predictions with a short time lag.

\begin{table}
\begin{center}
\begin{tabular}{|p{25mm}|c|c|c|c|c|}
\hline
\begin{tabular}{@{}c@{}}Target width, \\n frames\end{tabular}  & 1 & 3 & 5 & 7 & 9 \\
\hline\hline
PCE & 0.947 & 0.940 & 0.965 & 0.970 & \textbf{0.979}\\
SPCE & 0.931 & 0.923 & 0.954 & 0.965 & \textbf{0.975}\\
Global RMSE, px & 23.65 & 7.02 & 5.81 & 5.79 & \textbf{5.27}\\
Local RMSE, px & 5.53 & 2.27 & \textbf{1.71} & 2.40 & 2.03\\
Global accuracy & 0.962 & 0.978 & 0.981 & 0.979 & \textbf{0.981}\\
Local accuracy & 0.928 & 0.979 & \textbf{0.983} & 0.981 & 0.982\\
\hline
\end{tabular}
\end{center}
\caption{The influence of the length of the input sequence on the event spotting metrics. As the task is paired with ball detection, the results of the last are shown as well. Longer input sequences are beneficial for the event spotting.}
\label{tab:Len_events}
\end{table}

\subsection{Loss balancing}
Different ways of the TTNet loss components balancing were investigated to chose an optimal training approach. As a final loss, it was either just a sum of all components, or manually tuned constant weights for the loss components, or adaptively balanced loss described above. Expectedly, the latter method demonstrated the most promising results (see Tab. \ref{tab:Loss_compare}), giving the best metric values in more tasks and, most importantly, in the event spotting task.

\begin{table}
\begin{center}
\begin{tabular}{|p{26mm}|c|c|c|}
\hline
Loss & Unbalanced & \begin{tabular}{@{}c@{}}Manually \\ weighted\end{tabular} & Balanced \\
\hline\hline
Global RMSE, px & 9.34 & 7.74 & \textbf{6.79} \\
Local RMSE, px & 2.93 & 2.38 & \textbf{1.97} \\
Global accuracy & 0.977 & \textbf{0.977} & 0.975 \\
Local accuracy & 0.975 & 0.978 & \textbf{0.978} \\
IoU & \textbf{0.938} & 0.902 & 0.928 \\
PCE & 0.976 & 0.968 & \textbf{0.977} \\
SPCE & 0.966 & 0.963 & \textbf{0.970} \\
\hline
\end{tabular}
\end{center}
\caption{Comparison of training results of the TTNet with different loss aggregation strategies. The adaptive balancing performs better on most tasks and metrics}
\label{tab:Loss_compare}
\end{table}
\section{Inference on a live video stream}
As it was mentioned above, the predictions of the proposed neural network should be supplemented with the serve event spotting approach. It was done with a separate convolutional neural network, which adopts architecture and the training method from the event spotting branch of the multi-task neural network.

The game rules module, working on the raw predictions, was developed to enable completely automated table tennis referring. The whole system demo video is provided in supplementary materials. It demonstrates the performance of the neural network for the purposes of the auto-referee system.

\section{Conclusion}
This paper introduces a lightweight multi-task architecture TTNet for real-time data extraction from table tennis videos. It works on downscaled full HD videos and is able to detect the ball with pixel-level accuracy, using a cascade of two detectors, working on different resolutions; spots fast in-game events and predicts semantic segmentation masks while processing 120 fps with a single consumer-grade GPU. Additionally, we demonstrated that all branches of the proposed neural network could be trained simultaneously in an end-to-end manner. Thereby, our work contributes to the development of deep learning-based approaches for sports analysis. We also publish a labeled multi-task dataset for development and assessment of computer vision-based systems of table tennis analysis.

\newpage

{\small
\bibliographystyle{bibtex/ieee_fullname}
\bibliography{bibtex/egbib}
}

\end{document}